\documentclass[a4paper]{article}

\usepackage{INTERSPEECH2018}
\usepackage{subcaption}
\usepackage{cite}
\usepackage[table]{xcolor}

\title{Building a Unified Code-Switching ASR System for South African Languages}
\name{Emre Y\i lmaz$^{1,2}$, Astik Biswas$^3$, Ewald van der Westhuizen$^3$, Febe de Wet$^3$ and Thomas Niesler$^3$\thanks{This work was performed while the first author was on a research visit to the Stellenbosch University.}}
\address{
  $^1$CLS/CLST, Radboud University, Nijmegen, Netherlands\\
  $^2$ Dept. of Electrical and Computer Engineering, National University of Singapore, Singapore \\
  $^3$Dept. of Electrical and Electronic Engineering, Stellenbosch University, South Africa}
\email{e.yilmaz@let.ru.nl, \{abiswas, ewaldvdw, fdw, trn\}@sun.ac.za}

\begin{document}

\maketitle
\begin{abstract}
We present our first efforts towards building a single multilingual automatic speech recognition (ASR) system that can process code-switching (CS) speech in five languages spoken within the same population. This contrasts with related prior work which focuses on the recognition of CS speech in bilingual scenarios. Recently, we have compiled a small five-language corpus of South African soap opera speech which contains examples of CS between 5 languages occurring in various contexts such as using English as the matrix language and switching to other indigenous languages. The ASR system presented in this work is trained on 4 corpora containing English-isiZulu, English-isiXhosa, English-Setswana and English-Sesotho CS speech. The interpolation of multiple language models trained on these language pairs enables the ASR system to hypothesize mixed word sequences from these 5 languages. We evaluate various state-of-the-art acoustic models trained on this 5-lingual training data and report ASR accuracy and language recognition performance on the development and test sets of the South African multilingual soap opera corpus.

\end{abstract}


\section{Introduction}
South Africa is a multilingual society with 11 official languages and since the majority of South Africans are multilingual, code-switching (CS) occurs commonly in everyday conversations. CS being a part of daily life, the phenomenon also commonly occurs in radio and TV broadcasts which makes broadcast archives valuable sources of CS speech data. We have recently compiled a CS speech corpus which contains 14.3 hours of language-balanced speech compiled from soap opera broadcasts. Our research focuses on designing the acoustic and language models that can operate on this type of multilingual CS speech.

The impact of CS and other kinds of language switches on the performance of speech-to-text systems has recently received research interest, resulting in several robust acoustic modeling \cite{stemmer2001,lyu2006,vu2012,modipa2013,
lyudovyk2014,wu2014,westhuizen2016,yilmaz2016_2,yilmaz2016_4} and language modeling \cite{li2012,adel2013,adel2014,zeng2017,hamed2017,westhuizen2017} approaches for CS speech. In previous work, the Radboud team has explored the ASR and code-switching detection performance of various acoustic models applied to CS Frisian-Dutch speech \cite{yilmaz2017_1,yilmaz2016_2,yilmaz2016_4}. We further proposed several ways of increasing the amount of available training speech data by applying several automatic transcription strategies \cite{yilmaz2017_2,yilmaz2018}. 

Meanwhile, the Stellenbosch team has been developing a CS ASR system for isiZulu-English CS speech with a focus on language modeling~\cite{westhuizen2016,westhuizen2017}. The improvements in this system obtained by using speech data from different CS language pairs are explored in another submission to this conference \cite{biswas2018}.

In this work, we describe our joint efforts to build a 5-lingual ASR system that can recognize all five languages present in the South African Soap Opera Corpus, namely English, isiZulu, isiXhosa, Setswana and Sesotho. For this purpose, we develop acoustic and language models that enable language switches between these five languages. Unlike the prior work focusing on bilingual CS scenarios, this ASR system can hypothesize CS word sequences containing all five languages. 

We first use monolingual and bilingual text to train a language model with words from all target languages. The only source of CS text is the transcriptions of the training speech data which contain 156k words in total. Then, we train several recently proposed acoustic models on the four different CS pairs present in the corpus and apply these to the combination of all development and test data. We report 5-lingual ASR accuracies and language recognition (LR) performance of the developed ASR system.

This paper is organized as follows. Section~\ref{sec:lang} introduces the demographics and the linguistic properties of the target South African languages. Section~\ref{sec:cor} summarizes the South African Soap Opera Corpus that has recently been collected for CS speech research. Section~\ref{sec:asr} describes the details of the 5-lingual CS ASR system. The experimental setup is described in Section~\ref{sec:exps} and the ASR and LR performance of the described ASR system is presented in Section~\ref{sec:res}. Section~\ref{sec:conc} concludes the paper.
 
\section{Target South African Languages}
\label{sec:lang}
The linguistic and phonetic properties of the target indigenous languages are given in the following paragraphs. All four languages belong to Southern Bantu language family. isiZulu and isiXhosa are Nguni languages and linguistically similar. Furthermore, Sesotho and Setswana belong to the Sotho family and are also linguistically similar. All are agglutinative, tonal and click languages written in the Latin alphabet. The information presented in the coming paragraphs is extracted from the Ethnologue\footnote{https://www.ethnologue.com/}, UCLA Language Materials Project\footnote{http://www.lmp.ucla.edu/} and Census documents\footnote{http://www.statssa.gov.za}. We refer the reader to these sources (and the references therein) for further details.

The isiZulu language has 11.5M native and 15.5M second language (L2) speakers mostly living in South Africa. As many other indigenous languages in South Africa, it has relatively recently become a written language. The Zulu phonology is characterized by a simple vowel inventory with 5 vowels and a highly marked consonantal system with ejectives, implosives and clicks~\cite{zulu}. Zulu has borrowed many words from other languages, especially Afrikaans and English.

The second language, isiXhosa, has 8M native and 11M L2 speakers mostly living in South Africa. IsiXhosa has 58 consonants including 18 click consonants, 10 vowels and two tones. It is historically related to the Khoisan Languages, i.e. languages of southern Africa hunter-gatherer populations, and it has borrowed many words from these languages and later from English and Afrikaans.

Thirdly, Sesotho is spoken by 6M native and 8M L2 speakers in South Africa and Lesotho. It has 9 vowels and 39 consonants including ejectives, clicks and uvular trill. Various sound changes are observed involving vowels and consonants including various sorts of assimilation, elision, vowel merging and devoicing~\cite{sesotho}. The fourth and the final language, Setswana, is spoken by 5M native and 7.5M L2 speakers in South Africa and Botswana. It includes 7 vowels and 29 consonants, 3 of which contain clicks. There are two tones which are orthographically not marked. Despite a high mutual intelligibility with Sesotho, they are generally considered to be two separate languages.

\begin{table}[t]
	\centering
	\caption{Duration of English, isiZulu, isiXhosa, Setswana, Sesotho monolingual (emdur, zmdur, xmdur, tmdur, smdur)  and CS (ecsdur, zcsdur, xcsdur, tcsdur, scsdur) utterances~\cite{westhuizen2018}}
	\label{tab:stat}
	\resizebox{\columnwidth}{!}{%
		\begin{tabular}{|c|r|r|r|r|r|}
			\hline
			\multicolumn{6}{|c|}{English-isiZulu} \\ \hline \hline
			\textbf{Set} & \textbf{emdur} & \textbf{zmdur} & \textbf{ecsdur} & \textbf{zcsdur} & \textbf{total} \\ \hline
			\textbf{Train} & 1.55h & 1.55h & 45.86m & 56.99m & 4.81h \\ \hline
			\textbf{Dev} & - & - & 4.01m & 3.96m & 8m \\ \hline
			\textbf{Test} & - & - & 12.76m & 17.85m & 30.4m \\ \hline
			\textbf{Total} & 1.55h & 1.55h & 1.04h & 1.31h & 5.45h \\ \hline \hline
			\multicolumn{6}{|c|}{English-isiXhosa} \\ \hline \hline
			\textbf{Set} & \textbf{emdur} & \textbf{xmdur} & \textbf{ecsdur} & \textbf{xcsdur} & \textbf{total} \\ \hline
			\textbf{Train} & 65.22m & 53.55m & 18.04m & 23.73m & 160.54m \\ \hline
			\textbf{Dev} & 2.86m & 6.48m & 2.21m & 2.13m & 13.68m \\ \hline
			\textbf{Test} & - & - & 5.56m & 8.78m & 14.34m \\ \hline
			\textbf{Total} & 68.08m & 60.03m & 25.81m & 34.64m & 3.143h \\ \hline \hline
			\multicolumn{6}{|c|}{English-Setswana} \\ \hline \hline
			\textbf{Set} & \textbf{emdur} & \textbf{tmdur} & \textbf{ecsdur} & \textbf{tcsdur} & \textbf{total} \\ \hline
			\textbf{Train} & 40.4m & 30.96m & 34.37m & 34.01m & 139.74m \\ \hline
			\textbf{Dev} & 0.76m & 4.26m & 4.54m & 4.27m & 13.83m \\ \hline
			\textbf{Test} & - & - & 8.87m & 8.96m & 17.83m \\ \hline
			\textbf{Total} & 41.16m & 35.22m & 47.78m & 47.24m & 2.86h \\ \hline \hline
			\multicolumn{6}{|c|}{English-Sesotho} \\ \hline \hline
			\textbf{Set} & \textbf{emdur} & \textbf{smdur} & \textbf{ecsdur} & \textbf{scsdur} & \textbf{total} \\ \hline			\textbf{Train} & 49.34m & 35.32m & 23.02m & 34.04m & \multicolumn{1}{r|}{141.72m} \\ \hline
			\textbf{Dev} & 1.09m & 5.05m & 3.03m & 3.59m & \multicolumn{1}{r|}{12.77m} \\ \hline
			\textbf{Test} & - & - & 7.80m & 7.74m & \multicolumn{1}{r|}{15.54m} \\ \hline
			\textbf{Total} & 50.43m & 40.37m & 33.85m & 45.37m & \multicolumn{1}{r|}{2.83h} \\ \hline
		\end{tabular}%
	}
\end{table}

\section{South African Soap Opera Corpus}
\label{sec:cor}
A multilingual corpus containing examples of CS speech has recently been compiled from 626 South African soap opera episodes. The ELAN media annotation tool~\cite{wittenburg2006} has been used to segment and annotate the data. The spontaneous nature of the speech and the presence of various CS types makes this type of speech interesting for designing an ASR system which is expected to operate on CS speech from South Africa.

The corpus is still under development and the version we used corresponds to the language-balanced dataset with 14.3 hours of speech introduced in~\cite{westhuizen2018}. The data contains examples of CS between South African English, isiZulu, isiXhosa, Setswana and Sesotho. The corresponding code-switch language pairs are referred to as English-isiZulu, English-isiXhosa, English-Setswana, and English-Sesotho. An overview of the statistics for the training (Train), development (Dev) and test (Test) sets for each language pair is given in Table~\ref{tab:stat}. Each data set is described in terms of its total duration as well as the duration of the monolingual (m) and CS (cs) segments.

The soap opera speech is typically fast, spontaneous and may express emotion, with a speech rate that is between 1.22 and 1.83 times higher than prompted speech in the same languages. Among the 10\,343 code-switched utterances in the corpus, 19\,207 intra-sentential language switches are observed. Insertional code-switching with English words is observed to be most frequent. Intra-word CS occurring when English words are supplemented with Bantu affixes in an effort to conform to Bantu phonology is also observed. Note that the test utterances always contain CS and are never monolingual.

\section{5-lingual CS ASR System}
\label{sec:asr}

With the ultimate goal of an ASR system that can operate on all South African languages and correctly process the language switches, we build a language and an acoustic model that considers the word inventory of the five languages of our corpus. We apply the ideas that have been shown to be useful in monolingual scenarios to our multilingual CS system to determine the ASR performance that can be obtained using modern acoustic and language models.

For language modeling, both monolingual and CS text resources were used in varying quantities based on availability. CS text is practically non-existent and CS in textual resources hardly occurs. As is common, we use the transcriptions of the CS speech data for language model training purposes. Since the CS text constitutes a relatively small component of all available text data, we merge all available CS text to train a 5-lingual CS language model. Secondly, all monolingual text from the African languages is merged to train a 4-lingual LM. These models are later interpolated with the English monolingual model that has been trained on a much larger monolingual corpus. Following this strategy provided the lowest perplexities on the transcriptions of the development data in pilot experiments.

For training the acoustic models, we have only used the balanced corpora without including any other available monolingual corpora. On this small amount of data, we explore the performance of various NN architectures including conventional fully connected DNNs, time-delay NN (TDNN)~\cite{waibel1989,peddinti2015} and its combination with recurrent NN architectures such as long short-term memory (LSTM) and bidirectional LSTM~\cite{peddinti2017} using different acoustic features. Since most of the target languages are tonal, we also extract and attach pitch information to the acoustic features.

Together with the ASR performance, we also evaluate the LR accuracies of the described ASR systems to gain an insight into how well it can distinguish between the target languages, especially between those most similar (isiZulu-isiXhosa and Setswana-Sesotho). To achieve this, we analyze the confusion in the language tags assigned to each word during recognition. 

\section{Experimental Setup}
\label{sec:exps}

\begin{table}
\centering
\caption{The total number of words in each subcorpus used for LM training}
\addtolength{\tabcolsep}{-1.3pt}
\begin{tabular}{| l | r | l | r |}
\hline
\multicolumn{2}{|c|}{CS} & \multicolumn{2}{c|}{Monolingual} \\
\hline
Language pairs & \# of Words & Language & \# of Words \\
\hline \hline
English-isiZulu & 52k  & English & 470M \\
\hline
English-isiXhosa & 32k & isiZulu & 3.2M \\
\hline
English-Setswana & 35k & isiXhosa & 1.4M \\
\hline
English-Sesotho & 35k  & Setswana & 2.8M \\
\hline
All CS text & 156k & Sesotho & 0.2M \\
\hline
\end{tabular}
\label{tab:lm_data}
\end{table}

\subsection{Language Modeling}

The available monolingual and CS text in total number of words is presented in Table~\ref{tab:lm_data}. The texts used for monolingual LM training were collected from various sources which include online newspapers, magazines and newsletters (South African English, isiZulu, isiXhosa, Setswana), web text from the Leipzig Corpus Collection~\cite{biemann2007} (South African English, isiZulu, isiXhosa, Sesotho), parliamentary bulletins (isiXhosa, Setswana), and the Babel corpus transcriptions (isiZulu).

The language models used in these experiments are 5-lingual 3-gram and 5-gram with interpolated Kneser-Ney smoothing~\cite{kneser1995} for recognition and lattice rescoring respectively. We interpolate: (1) a CS 3-gram trained on all CS text, (2) a 4-lingual 3-gram trained on all monolingual text from 4 African languages, and (3) an English 3-gram to obtain the final 3-gram LM. The interpolation weights are learned on the transcriptions of the development data. We have observed that assigning relatively higher weights (0.85-0.9) to the CS LM reduces the perplexities considerably. The final 3-gram model has perplexities of 412 and 617 on the development and test transcriptions respectively. For the final 5-gram model, the corresponding figures are 402 and 605.

\subsection{Acoustic Modeling}

The recognition experiments are performed using the Kaldi ASR toolkit~\cite{kaldi}. We train a conventional context dependent Gaussian mixture model-hidden Markov model (GMM-HMM) system with 25k Gaussians using 39 dimensional mel-frequency cepstral coefficient (MFCC) features including the deltas and delta-deltas to obtain the alignments for training the NN models. 

As a reference, DNNs with 6 hidden layers and 2048 sigmoid hidden units at each hidden layer are trained on the 40-dimensional log-mel filterbank (FBANK) features with the deltas and delta-deltas. DNN training is performed by mini-batch Stochastic Gradient Descent with an initial learning rate of 0.008 and a minibatch size of 256. The time context size is 11 frames achieved by concatenating $\pm$5 frames. We further apply sequence training using a state-level minimum Bayes risk (sMBR) criterion~\cite{vesely2013}.

In addition, we train TDNN (3 standard, 6 time-delay layers), TDNN-LSTM (1 standard, 6 time-delay and 3 LSTM layers) and TDNN-BLSTM (1 standard, 2 time-delay and 3 BLSTM layers) acoustic models with the lattice-free maximum mutual information (LF-MMI) criterion~\cite{povey2016} according to the standard recipe provided for the Switchboard database in the Kaldi toolkit (ver.~5.2.99). With these models, we use 40-dimensional MFCCs together with 3-dimensional pitch features (appended when mentioned). The training parameters provided in the recipe are used without performing any parameter tuning. The 3-fold data augmentation~\cite{ko2015} is applied to the training data. 

\subsection{Pronunciation Lexicon}

The pronunciation lexicon contains 23\,453 words of which 5965 are English, 7448 are isiZulu, 5975 are isiXhosa, 1625 are Setswana and 2437 are Sesotho. Due to the pronunciation variants, the lexicon has 30\,489 entries in total. The phonetic alphabet contains a total of 284 phones of which 45 are English, 49 are isiZulu, 66 are isiXhosa, 59 are Setswana and 65 are Sesotho. The ASR experiments are closed vocabulary implying that there are no out-of-vocabulary words in the development and test data. 

\section{Results and Discussion}
\label{sec:res}

We present the results of the ASR experiments in this section. The ASR quality is quantified by calculating the word error rate (WER) with the language tags of the words removed. The language recognition performance is later evaluated in the form of a confusion matrix with an extra row and column to accommodate insertion and deletion errors. 

\begin{table}
\centering
\caption{WER (\%) on development and test sets provided by different acoustic models}
\addtolength{\tabcolsep}{-4pt}
\begin{tabular}{| l | c | c | c | c | c |}
\hline
AM & Features  & LM        &  Dev     & Test & Total \\
\hline \hline
DNN & 40-FBANK    & 3G        &   65.8   &   66.9 &  66.5 \\
\hline
TDNN & 40-MFCC    & 3G        &   61.2   &   60.2 &  60.6 \\
\hline
TDNN+LSTM & 40-MFCC & 3G  &  59.5    &   58.3     &  58.8 \\
\hline
TDNN+BLSTM & 40-MFCC& 3G  &  57.0    &   57.3     &  57.2 \\
\hline \hline
TDNN+BLSTM & 40-MFCC+Pitch & 3G &  55.9  &  56.2  &  56.1 \\
\hline
TDNN+BLSTM & 40-MFCC+Pitch & 3G+5G & \bf{55.5} & \bf{55.7} & \bf{55.6} \\
\hline
\end{tabular}
\label{tab:ASR_1}
\end{table}

\begin{table*}
\centering
\caption{WER (\%) on the development and test sets of each CS subcorpus in provided by the best recognition system presented in Table~\ref{tab:ASR_1} - The total number of words in each subcorpus is given in parentheses.}
\begin{tabular}{| l | c | c | c | c | c | c |}
\hline
CS Pair ($L_{1}$-$L_{2}$) & \multicolumn{2}{c|}{$L_{1}$-$L_{2}$} & \multicolumn{2}{c|}{$L_{1}$}  &  \multicolumn{2}{c|}{$L_{2}$} \\
\hline 
                 & Dev & Test& Dev & Test& Dev & Test \\
\hline
English-isiZulu  &  44.2 (1572)& 52.6 (5658)& 33.7\,\,\,\,(838)& 38.6 (2459)& 56.1\,\,\,\,(734)& 63.3 (3199)  \\
\hline
English-isiXhosa &  51.9 (2300)& 62.8 (2651)& 39.6 (1153)& 44.9 (1149)& 64.3 (1147)& 76.5 (1502)  \\
\hline
English-Setswana &  56.7 (3707)& 52.3 (4939)& 38.0 (1170)& 33.9 (1970)& 65.4 (2537)& 64.6 (2969)  \\
\hline
English-Sesotho  &  62.6 (3067)& 59.1 (4054)& 41.5\,\,\,\,(843)& 41.9 (1794)& 70.6 (2224)& 72.8 (2260)  \\
\hline
\end{tabular}
\label{tab:ASR_2}
\end{table*}

\subsection{ASR Results}

First, we investigate the best performing NN architecture and then we move to a detailed analysis of the recognition accuracy on each language component. The ASR results obtained on the development and test sets using different acoustic models are presented in Table~\ref{tab:ASR_1}. 

A conventional fully connected DNN provides a total WER of 66.5\%, while using a TDNN model brings considerable improvements reducing the WER to 60.6\%. Adding recurrent layers helps by further reducing the total WER to 58.8\%. With a WER of 57.2\%, the ASR system using bidirectional recurrent layers together with time-delay layers has the lowest WER among all the aforementioned acoustic models.

We further explore the impact of appending pitch features and lattice rescoring using a larger n-gram language model. Using pitch features is common practice when building ASR systems for tonal languages. In this scenario, using pitch features provides an absolute improvement of 1.7\% leading to a WER of 56.1\%. 5-gram lattice rescoring brings further marginal improvements reducing the total WER to 55.6\%. In the remaining experiments, we analyze the output of this ASR system to gain better insight to language-specific ASR performance and language recognition performance.

The language-specific recognition accuracies for each CS pair and dataset are shown in Table~\ref{tab:ASR_2}. From these results, it can be seen that there is a large performance gap between the general ASR performance on English and the African languages. English being spoken in all 4 CS pairs, the amount of English training data is much larger than it is for the other languages. 

A second issue to be addressed is the poor performance for Sesotho. Even though Sesotho has a similar amount of acoustic training data as Setswana, there is very little textual data available in this language for LM training (0.2M words compared to the 2.8M of Setswana). This results in WER that are higher than 70\%, while the WER for the English words is approximately 42\%. 

A final observation is the similar performance on the development and test data of the Sotho languages (Setswana-Sesotho) implying that the ASR performance on monolingual segments and segments with CS are rather similar. It is worth remarking that the utterances in the test sets all contain CS, while some utterances in the development data (except for English-isiZulu) are monolingual which are expected to be easier to recognize. However, we only observe this pattern in isiXhosa where there is a performance gap between the development and test sets in contrast to the Sotho languages. Having no monolingual utterances in both sets and a large difference in development and test set size (which may result in larger variance in WERs), we do not consider English-isiZulu results in this comparison. 

\begin{table}
\caption{Confusion matrices of the hypothesized language tags on the development and test data}
\vspace{-0.2cm}
\begin{subtable}{.46\textwidth}
\centering
\caption{Dev \label{tab:lr_dev}}
\begin{tabular}{| l || c | c | c | c | c | c |}
\hline
       & ENG     & ZUL  &  XHO    & TSN  &   SOT   &  DEL \\
\hline \hline
ENG    &3336   &  152   &   25    &  87  &    72   &  327 \\
\hline
ZUL    &  84   &   \cellcolor{blue!25}513   &     \cellcolor{blue!25}10    &  11  &    17   &  99 \\
\hline
XHO    & 166   &   \cellcolor{blue!25}313   &   \cellcolor{blue!25}531    &  25  &    23   &  89 \\
\hline
TSN    & 373   &  153   &   20    &\cellcolor{green!25}832  &   \cellcolor{green!25}626   &  535 \\
\hline
SOT    & 361   &  209   &    11    & \cellcolor{green!25}528  &   \cellcolor{green!25}732   &  386 \\
\hline
INS    & 159   &   57   &   8    &  52  &    63   &    0 \\
\hline
\end{tabular}
\vspace{0.2cm}
\end{subtable}
\begin{subtable}{.46\textwidth}
\centering
\caption{Test \label{tab:lr_test}}
\begin{tabular}{| l || c | c | c | c | c | c |}
\hline
       & ENG     & ZUL  &  XHO    & TSN  &   SOT   &  DEL \\
\hline \hline
ENG    &5953   &  309   &   62    &  174  &    165   &  694 \\
\hline
ZUL    &  474   &  \cellcolor{blue!25}2010   &    \cellcolor{blue!25}64    &  88  &    120   &  443 \\
\hline
XHO    & 262   &  \cellcolor{blue!25}496   &  \cellcolor{blue!25}447    &  35  &    41   &  221 \\
\hline
TSN    & 411   &  138   &   20    &\cellcolor{green!25}875  &   \cellcolor{green!25}810   &  725 \\
\hline
SOT    & 398   &  180   &   20    & \cellcolor{green!25}680  &   \cellcolor{green!25}457   &  530 \\
\hline
INS    & 211   &   77   &   16    &  62  &    56   &    0 \\
\hline
\end{tabular}
\end{subtable}
\label{tab:LR_1}
\end{table}

\subsection{LR Results}

Using the language tags assigned to each word by the best-performing ASR system in Table~\ref{tab:ASR_1} for the development and test sets, the confusion matrices shown in Table~\ref{tab:LR_1} are obtained. Confusions between the isiZulu-isiXhosa and Setswana-Sesotho language pairs are marked in purple and green respectively. This is done to highlight the cells where higher confusion is expected due to the similarity between the two languages.

Focusing firstly on isiZulu-isiXhosa, we see that the confusion occurs mostly in a single direction, i.e. many more isiXhosa words are identified as isiZulu words. In the second language pair, Setswana-Sesotho, the confusions occur in both directions in both development and test sets. The language recognition performance of the ASR system is significantly worse than any other language couple. This can be explained by the greater linguistic similarity between the languages and their larger intersection in the phoneme set and vocabulary. The lower acoustic and written resources further reduces the LR performance of the ASR system on the Sotho languages.

\section{Conclusion}
\label{sec:conc}

We present a first 5-lingual CS ASR system that is designed to recognize CS speech in 5 South African languages. Using a recently compiled soap opera speech corpus, we explore how well modern NN-based acoustic models can deal with the language switches given the limited availability of resources for the target languages. Language recognition performance implicit in the ASR is also evaluated, especially between the linguistically similar languages. We believe that these first findings are encouraging and provide insight into the challenges in building a unified CS system for multilingual countries such as South Africa.

\section{Acknowledgements}
\label{sec:acknow}
This research is funded by the NWO Project 314-99-119 (Frisian Audio Mining Enterprise). The authors would like to thank the Department of Arts \& Culture of the South African government for funding this research as well as Stellenbosch University for the travel grant that enabled the first author's visit to Stellenbosch.
\bibliographystyle{IEEEtran}

\bibliography{refs}

\end{document}